\documentclass{article}
\usepackage{graphicx} 

\usepackage{comment}
\usepackage{graphicx}
\usepackage{subfig}
\usepackage{longtable}
\usepackage{newtxtext,newtxmath}

\title{Automatic coral reef fish identification \\ and 3D measurement in the wild}
\author{Cyril Barrelet$^{1}$, Marc Chaumont$^{1,2}$, Gérard Subsol$^{1}$\\
1. Research-team ICAR, LIRMM, Univ Montpellier, CNRS, France, \\2. Univ Nîmes, France \\\{cyril.barrelet, marc.chaumont, gerard.subsol\}@lirmm.fr}
\date{May 2023}

\begin{document}

\maketitle

\section{Introduction}
Monitoring marine biodiversity has become essential to track potential species decline or invasive species due to overfishing, pollution, or global warming. \\
Marine ecologists used to dive underwater and count species on the fly, sometimes estimating their length by eye. As those missions are costly, Computer vision methods have become very promising in this field. While a person can dive for about one hour, multiple underwater cameras can capture video for several hours, resulting in thousands of videos. \\
Analyzing this material is very time-consuming, and therefore many works have been done to count fishes within videos automatically (\cite{counting}, \cite{counting3}, \cite{counting2}). \\
However, more than counting fish is needed to monitor marine life effectively. Indeed, another indicator is the inter-species biomass, which needs to know the fish species and their length. Acquiring this length is possible with stereoscopic cameras through a camera calibration process. \\
VidSync \cite{vidsync} is a software that helps marine ecologists calibrate cameras and measure fish manually within a video. An operator must select points in both left and right images and get a 3D measurement thanks to the calibration of the camera. \\
This method is tedious, considering the number of videos, and it is prone to error since the operator also has to identify the species. \\
Many works were interested in fish detection (\cite{villon:hal-01374123}, \cite{minutolo_fish_2020}, \cite{det_class}, \cite{wageeh_yolo_2021}), species identification (\cite{cabreira_artificial_2009}, \cite{fewshot}), and fish movement tracking (\cite{fishtrack}, \cite{trackfish2}, \cite{track_det}) on mono cameras, while a few worked on stereo cameras (\cite{track_stereo}, \cite{measurementfish}). None of them have since automatically detected, identified, measured, and tracked fish in wild underwater environments. \\
Thus, this paper introduces an automatic 3D fish population analysis pipeline within wild underwater videos.

\section{Materials and Methods}
\subsection{Method overview}

As our automatic 3D fish analysis procedure flowchart shows in Figure~\ref{chart}, the very first blocs consist in getting stereo videos. 

\begin{figure}[!htbp]
    \centering
    \includegraphics[width=8cm]{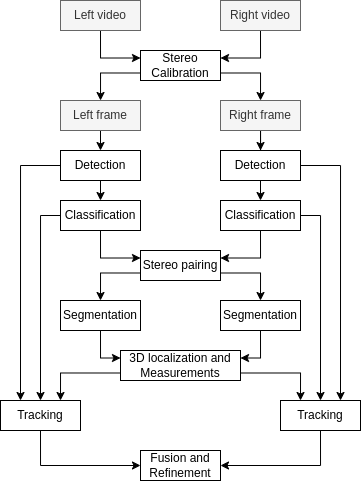}%
    \caption{\textbf{Flowchart} Our automatic 3D fish population analysis pipeline.}%
    \label{chart}
\end{figure}

\begin{figure}[!htbp]
    \centering
    \includegraphics[width=6cm]{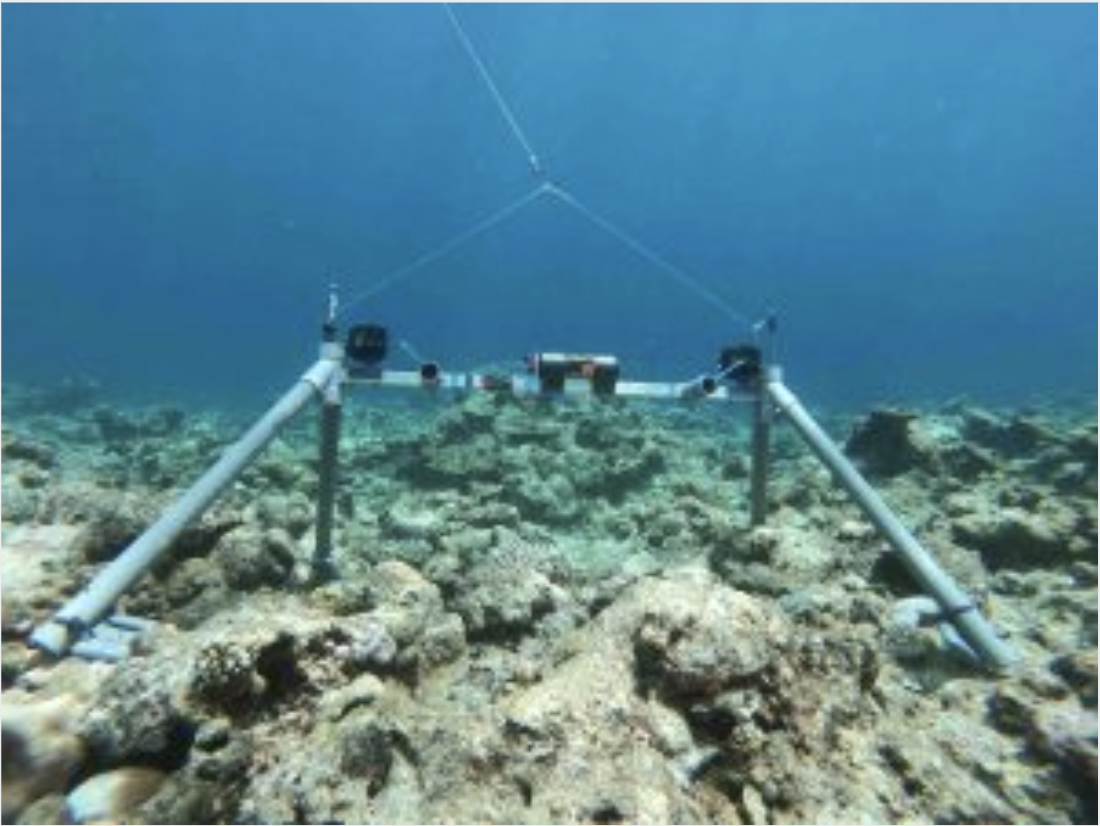}%
    \caption{\textbf{Our stereo system} Two GoPro Hero 4 cameras fixed on a rigid rig.}%
    \label{rig}
\end{figure}

Hence, we first recorded videos in various environments with a stereoscopic camera system, as shown in Figure~\ref{rig}, which consists of two GoPro\footnote{https://gopro.com} Hero 4 cameras with a resolution of 1080 $\times$ 1920, both housed in a diving case, fixed on a rigid rig 80 cm apart. Since both cameras need to be synchronized, we attached a blinking LED in front of each camera view, allowing for manual synchronization and preventing future divergence in the caption.

We summarize the pipeline as follows:

\begin{itemize}
   \item \textbf{Calibration}: From the left V\textsubscript{L} and right V\textsubscript{R} videos, we estimate the parameters of the cameras through a calibration procedure, detailed in section 2.2.
   
   \item \textbf{Detection}: Independently, each left I\textsubscript{L} and right I\textsubscript{R} image passes through our fish detection model, explained in section 2.3, acquiring the 2D localization of the detected fish through their corresponding bounding boxe BB in both images.
   
   \item \textbf{Classification}: The thumbnail of every detected left BB\textsubscript{L} and right BB\textsubscript{R} bounding box passes then through our species classification model, presented in section 2.4, identifying the species of every fish.

    \item \textbf{Stero pairing}: We rectify I\textsubscript{L} and right I\textsubscript{R} knowing the calibration parameters, allowing to pair, as described in section 2.5, each BB\textsubscript{L} to their corresponding BB\textsubscript{R}.

    \item \textbf{Segmentation}: Paired thumbnails pass through our fish segmentation model, presented in section 2.6, allowing us to create a pixel-wise mask of both left and right fish. 

    \item \textbf{3D localization and measurement}: Given the paired masks of every fish, we localize the 2D position of their snoot and tail and measure their corresponding 3D position, allowing us to get their size and 3D localization in the camera space, as explained in section 2.7.

    \item \textbf{Tracking}: We track the fish as long as they appear in the image, keeping their 2D and 3D positions and their top-5 species identification, as clarified in section 2.8.

    \item \textbf{Fusion and refinement}: Finally, we associate the same fish's left and right tracks, allowing for the smoothing of both the measurements and identification as described in section 2.9.

\end{itemize}

%
%
%
%
%
%

\subsection{Calibration}

\begin{figure}[!htbp]
    \centering
    \includegraphics[width=12cm]{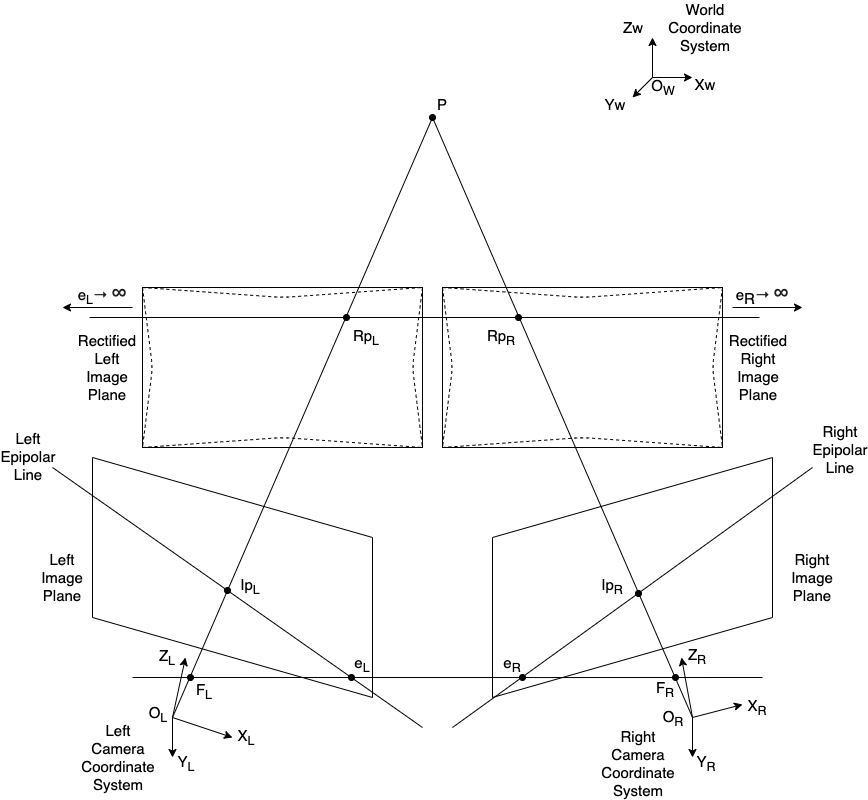}%
    \caption{\textbf{Epipolar geometry} Rectification scheme.}%
    \label{rectification}
\end{figure}

The camera calibration is a procedure allowing us to estimate cameras' parameters, distortion coefficients, and their relative position and orientation from one another. \\
By triangulation, it gives the ability to measure the distance between a point and the camera coordinate system noted \emph{O\textsubscript{L, R}} in Figure~\ref{rectification}.

\subsubsection*{METHOD}

After arriving at each dive location, the first step is to move a calibration pattern with known dimensions in front of the stereo cameras. 
As we will discuss later, different calibration pattern forms exist. Though, we used the calibration pattern generator from the OpenCV\footnote{OpenCV: https://opencv.org/} library \cite{opencv_library} for the sake of simplicity.\\
Given that the cameras are synchronized, meaning the first frame of one camera is temporally related to the first frame of the other one, the second step is to select 20~-~30 stereo frames, I\textsubscript{L}\textsuperscript{t\textsubscript{n}} and I\textsubscript{R}\textsuperscript{t\textsubscript{n}}, manually where both cameras can see the calibration pattern.\\
The final and automatic step is estimating, for each camera, their parameters, their distortion coefficient, and their relative position and orientation from one another.\\
The camera parameters form a so-called projection matrix of size $3\times4$, composed of the intrinsic and extrinsic parameters, as shown in (\ref{eu_eqn}).\\

\begin{equation} \label{eu_eqn}
\begin{bmatrix}
u\\v\\1
\end{bmatrix}
= \begin{bmatrix}
f\textsubscript{x} & 0 & c\textsubscript{x}\\ 
s & f\textsubscript{y} & c\textsubscript{y}\\ 
0 & 0 & 1
\end{bmatrix}
\begin{bmatrix}
r\textsubscript{11} & r\textsubscript{12} & r\textsubscript{13} & t\textsubscript{x}\\ 
r\textsubscript{21} & r\textsubscript{22} & r\textsubscript{23} & t\textsubscript{y}\\ 
r\textsubscript{31} & r\textsubscript{32} & r\textsubscript{33} & t\textsubscript{z}
\end{bmatrix}
\begin{bmatrix}
X\textsubscript{w}\\ 
Y\textsubscript{w}\\ 
Z\textsubscript{w}\\
1
\end{bmatrix}
\end{equation}

The intrinsic parameters link the pixel coordinate of an image point with the corresponding coordinate in the camera space. These consist of its focal length in mm, noted \emph{f\textsubscript{x}} and \emph{f\textsubscript{y}}, its optical center in pixel, noted \emph{c\textsubscript{x}} and \emph{c\textsubscript{y}}, and its skew coefficient, noted \emph{s}, defining the angle between the x and y pixel axes.\\
The extrinsic parameters describe the position and orientation of the camera in the world. These consist of a rotation matrix of size $3\times3$ noted \emph{r\textsubscript{ij}} and a translation matrix of size $3\times1$ noted \emph{t\textsubscript{x, y, z}}.\\
As shown in equation (1), the projection matrix allows transforming a 2D system coordinate, noted \emph{u} and \emph{v}, to a 3D world system coordinate noted \emph{X\textsubscript{w}}, \emph{Y\textsubscript{w}}, and \emph{Z\textsubscript{w}}.\\

Using those previously selected stereo frames and the OpenCV library implementation of Bouget's Calibration Toolbox\footnote{Bouget's Calibration Toolbox: http://www.vision.caltech.edu/bouguetj/calib} inspired by Zhang's technique \cite{zhang_flexible_2000}, we can automatically detect the calibration pattern corners, thanks to the Harris Corner Detector \cite{harris_combined_1988}, and extract the described camera parameters and the distortion coefficients for each camera. 
Then, we can estimate the relative position and orientation from the left to the right camera with the coordinates' corners set, the cameras' parameters, and the distortion coefficients.\\

\begin{figure}[!htbp]
    \centering
    \subfloat{{\includegraphics[width=13cm]{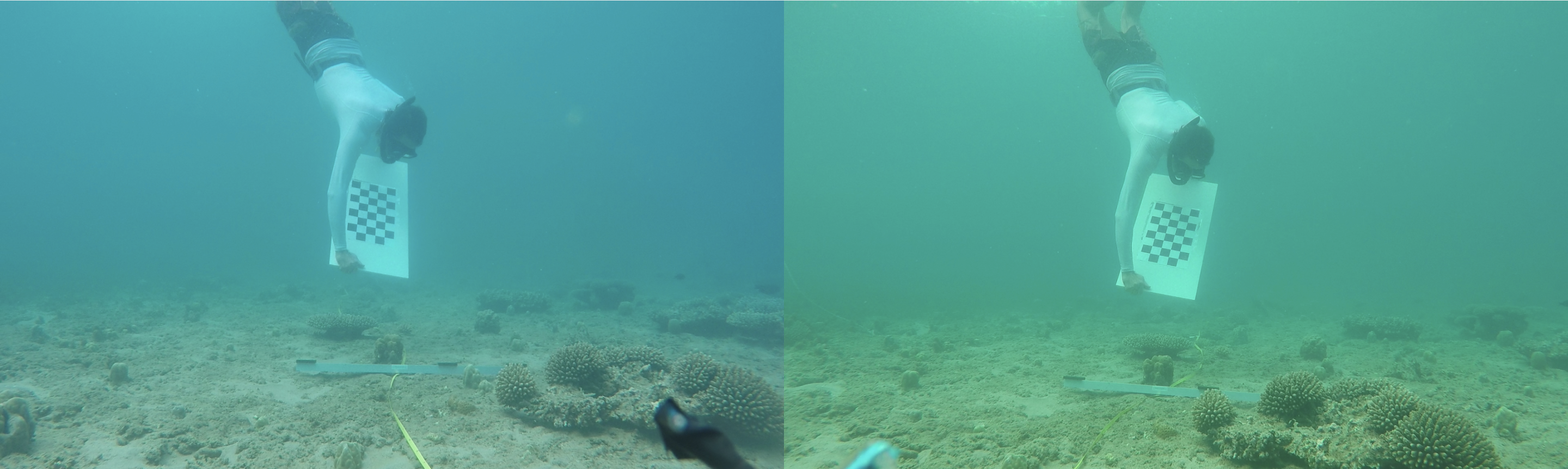}}}%
    \qquad
    \subfloat{{\includegraphics[width=13cm]{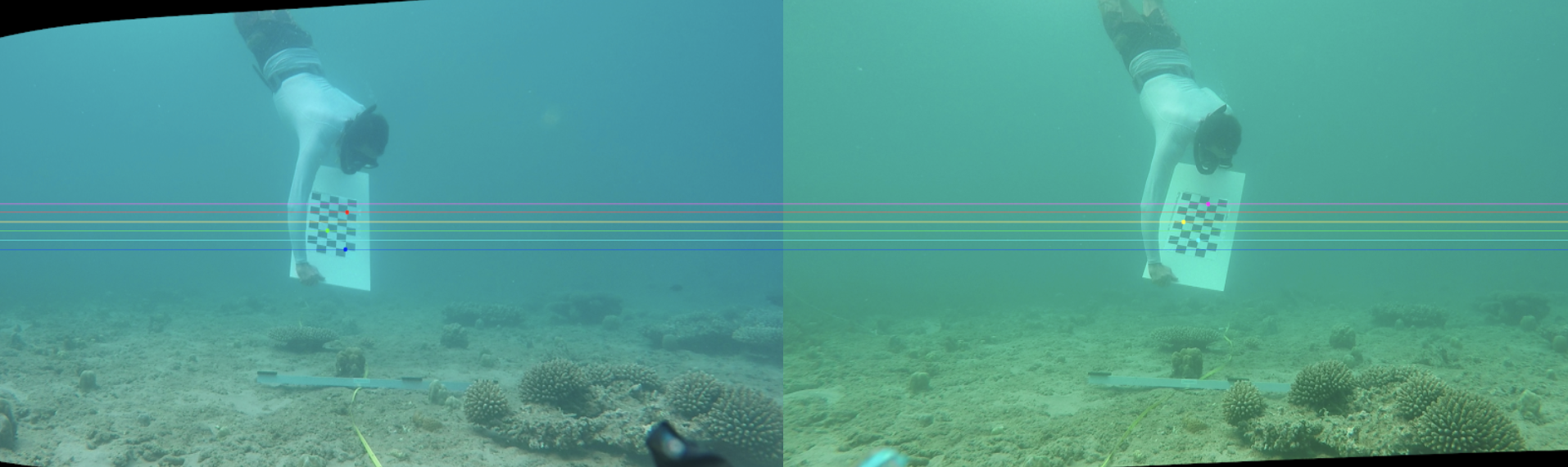}}}%
    \caption{\textbf{Image rectification} (Top) The original stereo frames with the calibration pattern. (Bottom) The same stereo frames but rectified.}%
    \label{recti_images}
\end{figure}

Given those calibration parameters, we correct the lens distortions and rectify the images. \\
As shown in Figure~\ref{recti_images}, the conjunction of the two focal points \emph{F\textsubscript{R}} and \emph{F\textsubscript{L}} of the right and left camera respectively intersects I\textsubscript{L} and I\textsubscript{R} in two specific points, \emph{e\textsubscript{L}} and \emph{e\textsubscript{R}}, called left and right epipoles, respectively \cite{zicari_efficient_2013}. 
If \emph{I\textsubscript{pR}} is the projection of the generic point \emph{P} of the scene into I\textsubscript{R}, its corresponding point \emph{I\textsubscript{pL}} in I\textsubscript{L} has to be searched for on the left epipolar line obtained connecting \emph{I\textsubscript{pL}} with the left epipole. 
The rectification is an image transformation that makes the epipolar lines collinear with the image scan lines; thus, the epipoles are shifted to infinity.\\
In other words, finding correspondence between each right \emph{R\textsubscript{pR}} rectified image pixel and its matching left \emph{R\textsubscript{pL}} rectified image pixel only requires searching in a unique axis along the epipolar lines, as shown in Figure~\ref{recti_images}. It is called the epipolar constraint. 

\subsubsection*{PRACTICAL CASE}

We printed a checkerboard with 5×8 tiles of 5~cm as the calibration pattern and plasticized it on a rigid board. \\
Then, we recorded a diver holding the board at various positions and orientations within 1~-~3~m from the stereo cameras rig. \\
Secondly, we manually synchronized  V\textsubscript{L} and V\textsubscript{R} thanks to the blinking LEDs and a clap that the diver did as a reference at the beginning of the recording. \\
Finally, we manually extracted 20-30 frames in which the calibration pattern could be detected in both I\textsubscript{L} and I\textsubscript{R}.\\
Then, we give those images to the calibration algorithm doing the rest automatically.

%
%
%
%
%
%

\subsection{Detection}

We train an object detection neural network to detect and localize fish within the images, as shown in the detection block of Figure~\ref{chart}. Many object detection neural networks exist: R-CNN \cite{girshick_rich_2014}, Fast R-CNN \cite{girshick_fast_2015}, YoloV3 \cite{redmon_yolov3_2018}, or YoloV4 \cite{bochkovskiy_yolov4_2020}, for instance. We chose YoloV5 \cite{glenn_jocher_2020_4154370} which is very efficient.\\
Once trained, the model estimates Bounding Boxes, BBs, at the fish location according to a threshold. This threshold allows retaining every BB whose confidence score, given by the regression, is not under it. We kept this threshold according to the original YOLOv5 implementation.\\
YOLOv5 can detect and identify an object simultaneously. However, it does not manage class balancing within the identification task, and since our fish detection database B\textsubscript{det} is poorly balanced, we get around this problem indirectly. 
Hence, we deliberately separate the localization from the identification task, allowing YOLOv5 to detect only fish-looking objects.

\subsubsection*{FISH DETECTION DATABASE}

Our B\textsubscript{det} consists of 3,362 manually annotated frames, with between 0 to 34 fish, extracted from various videos in shallow water taken in Mayotte and Scattered Islands at different times, for a total of 32,054 annotated BBs.\\
Additionally, every fish were identified. Every species has an average of 204 BBs but a standard deviation of 923 BBS because the number of BBs per species ranges from 1 to 10,675, as shown in Table~\ref{Bdet} (Supplementary materials).\\
We randomly split the database into a train, a validation, and a test set with a ratio of 80-10-10, respectively.

\subsubsection*{DATA AUGMENTATION}

Indeed, underwater images suffer from turbidity, color distortion, and low contrast because the light is attenuated while it propagates through water \cite{berman2020underwater}.\\
As more than the size of our database is needed to obviate those problems and to generalize the detection for every case, we use data augmentation techniques.\\
Thus, we performed color space and geometric transformations. We randomly apply a horizontal flip and modify the contrast within a range of 60 to 140\% \cite{shorten_survey_2019}. 

\subsubsection*{METHOD AND PARAMETERS}

\begin{figure}[!ht]
    \centering
    \includegraphics[width=10cm]{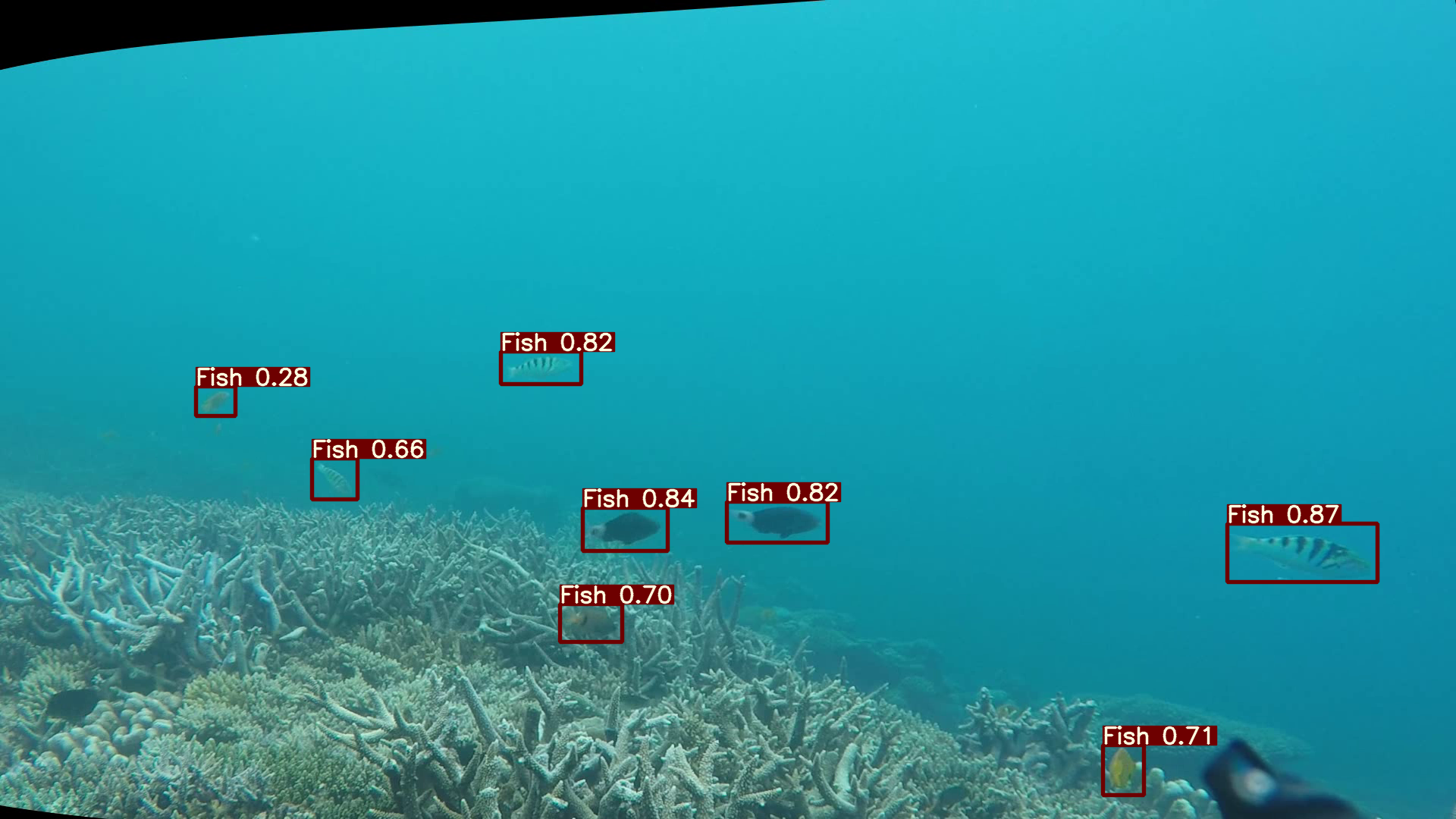}
    \caption{A frame where each fish-looking object is detected while specifying the confidence score YoloV5 attributes. }%
    \label{det}
\end{figure}

As we want YOLOv5 to detect only fish-looking objects, we set the number of classes to one. We chose the YOLOv5x6 model pre-trained on ImageNet at a resolution of 1280 × 1280 pixels. From its pre-trained weights, we used transfer learning to keep the previous knowledge. We chose the SGD optimizer and kept the same settings from the YOLOv5 V5.0 version. Finally, we set the batch size at four, the maximum according to our GPU capability (NVIDIA\footnote{https://www.nvidia.com/} Quadro RTX 6000 with 24Go VRAM).
We also investigated the batch size and resolution influence on different YOLOv5 models in the Supplementary material section. \\
Finally, we obtained a Precision of 82.9\% and a Recall of 84.7\%, with a mAP@.5 of 87.7\% on the test set.
Figure~\ref{det} shows the result of the detection.

%
%
%
%
%
%

\subsection{Classification}

The classification aims to train a Convolutional Neural Network for fish species identification. 
Once a fish is detected, its BB is cut out from the image. Then, this thumbnail passes through the neural network that gives us the logistic regression and hence its species and a confidence score. \\
As examples are needed to train a CNN, we used a species identification database. 

\subsubsection*{SPECIES IDENTIFICATION DATABASE}

Our species identification database B\textsubscript{id}  contains 56 different species classes and heights environmental classes such as algae or coral, as listed in Table~\ref{Bid} (Supplementary materials). 
Each class consists of (n=64) $2,757.6 \pm {1,844.8}$ thumbnails with diverse sizes manually extracted from various videos with the same conditions that our B\textsubscript{det} but at a different period, for a total of 44,625 annotated thumbnails.
As we did for the fish detection task, we split this database into three datasets respecting the same ratio. 

\subsubsection*{DATA AUGMENTATION}

To help generalize, we also used the same augmentation techniques mentioned before.\\
In addition, to prevent erroneous identification due to the occlusion problem within the fish detection task, we added 56 new classes representing species parts \cite{villon_deep_2018} to our database, which now has 120 different classes. Thus, we cut vertically and horizontally every species thumbnail while considering class balancing.

\subsubsection*{METHOD AND PARAMETERS}

\begin{figure}[!ht]
    \centering
    \includegraphics[width=10cm]{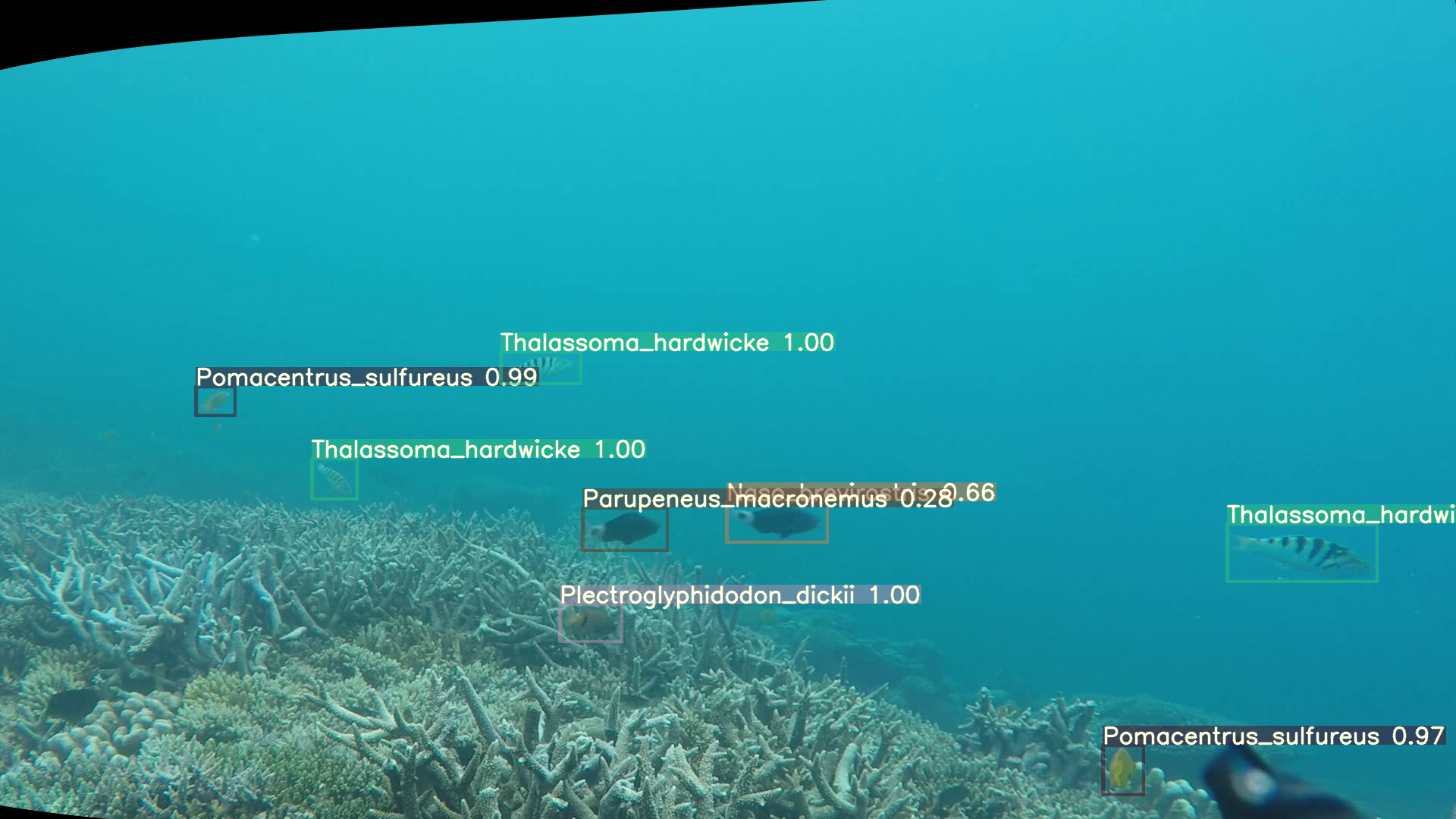}
    \caption{
    A frame where each BB is identified while specifying the confidence score EfficientNet attributes. The two central fish identified as \emph{Parupeneus macronemus} and \emph{Naso brevirostris} have a low confidence score, which suggests erroneous identification. This species is, in fact, not part of B\textsubscript{id} }%
    \label{class}
\end{figure}

We selected the EfficientNet-B6 model \cite{tan_efficientnet_2020}, which has 46M parameters, with slightly different settings from the original paper: ADAM optimizer with decay 0.9; batch norm momentum 0.99; weight decay $10^{-5}$
; dropout ratio 0.5; initial learning rate $10^{-3}$
; the cross-entropy loss function; and a batch size of 16. \\
Since species identification is challenging, we adopt a two-stage learning approach \cite{howard_universal_2018}. 
First, we only trained its classifier using transfer learning from the ImageNET pre-trained model while freezing all other layers. 
We dynamically decreased the initial learning rate by 0.1 whenever the validation loss plateaus according to a 'patience' number of epochs. \\
Then, using fine-tuning, we unfroze all the layers and retrain the entire model. To retain previous knowledge and avoid forgetting during the fine-tuning, we dropped the initial learning rate to $10^{-4}$
, decreasing it throughout the training similarly. \\
Additionally, we set an Early Stopping for both phases, allowing to stop training when parameter updates no longer improve the validation set since large models cannot overfit \cite{ruiz_analysis_2020}.\\
We obtained a 97.95\% top-5 accuracy after the fine-tuning phase on the test set. Given that the accuracy is nearly perfect, we assumed a possible no mismatch within our B\textsubscript{id}. Thus, we further tested the model on our B\textsubscript{det}, as the two databases were created at different times. As expected, we obtained a slightly lower performance reaching a 95.3\% top-5 accuracy on B\textsubscript{det}.\\
The result of the classification is shown in Figure~\ref{class}.
Since we can detect and identify fish in the stereo pair, we propose a method to associate every BB from I\textsubscript{L}, BB\textsubscript{L}, to their corresponding BB on I\textsubscript{R}, BB\textsubscript{R}.

%
%
%
%
%
%

\subsection{Stereo pairing}

While we need to find correspondence between pixels from I\textsubscript{L} and I\textsubscript{R}, even though we only look on one axis thanks to the image's rectification, this pixel matching could be misleading. 
Indeed, as we need to go through every pixel across the epipolar line to find correspondence, faulty matching could occur, resulting in an incorrect measurement. 
Thus, we proposed a method for matching left and right BBs, allowing robust pixel-level matching.  

\subsubsection*{METHOD AND ALGORITHM}

\begin{figure}[!htbp]
    \centering
    \includegraphics[width=8cm, height=3cm,]{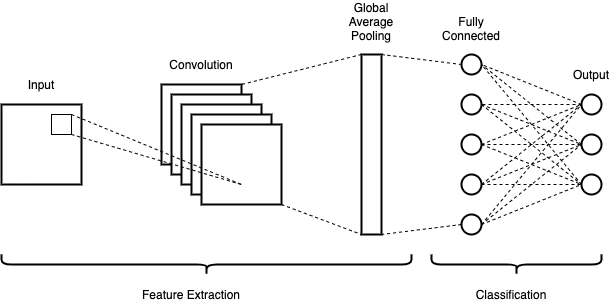}%
    \caption{Schematic diagram of a basic CNN}%
    \label{cnn}
\end{figure}

Firstly, for each BB\textsubscript{L} and BB\textsubscript{R}, we extract their $1\times2316$ dimension vector from the Global Average Pooling layer of the EfficientNet's feature extractor. Since this feature vector is previous to the classification network, as shown in Figure~\ref{cnn}, it is a thin-grained image representation of the species. Thereby, we can use it as a similarity metric.

\begin{figure}[!htbp]
    \centering
    \includegraphics[width=4cm, height=3cm]{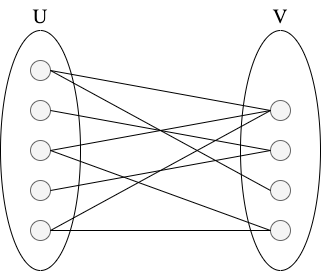}%
    \caption{\textbf{Bipartite graph:} BBs\textsubscript{L} IDs on the left and BBs\textsubscript{R} IDs with their respective cosine similarity as weights.}%
    \label{bipartite}
\end{figure}

Secondly, for each center coordinate of BBs\textsubscript{L}, we look for all center coordinates of BBs\textsubscript{R} along the same epipolar line within a delta of five pixels. \\
Moreover, we compute the Cosine similarity between their feature vector, giving an appearance distance. \\
Lastly, we define a bipartite graph consisting of two independent sets \emph{U} and \emph{V}, respectively BBs\textsubscript{L} and BBs\textsubscript{R} IDs. Then, we set the edges and weights accordingly to the previous steps, as shown in Figure~\ref{bipartite}.\\

Finally, we solve this one-to-one pairing optimization problem using the Hungarian algorithm \cite{kuhn_hungarian_1955}, maximizing the sum of the weights. 

Given that we do not have any available database to measure our method's performance yet, we tested it visually on a few stereo-pair images and assumed it works reasonably.

\begin{figure}[!htbp]
    \centering
    \includegraphics[width=12cm]{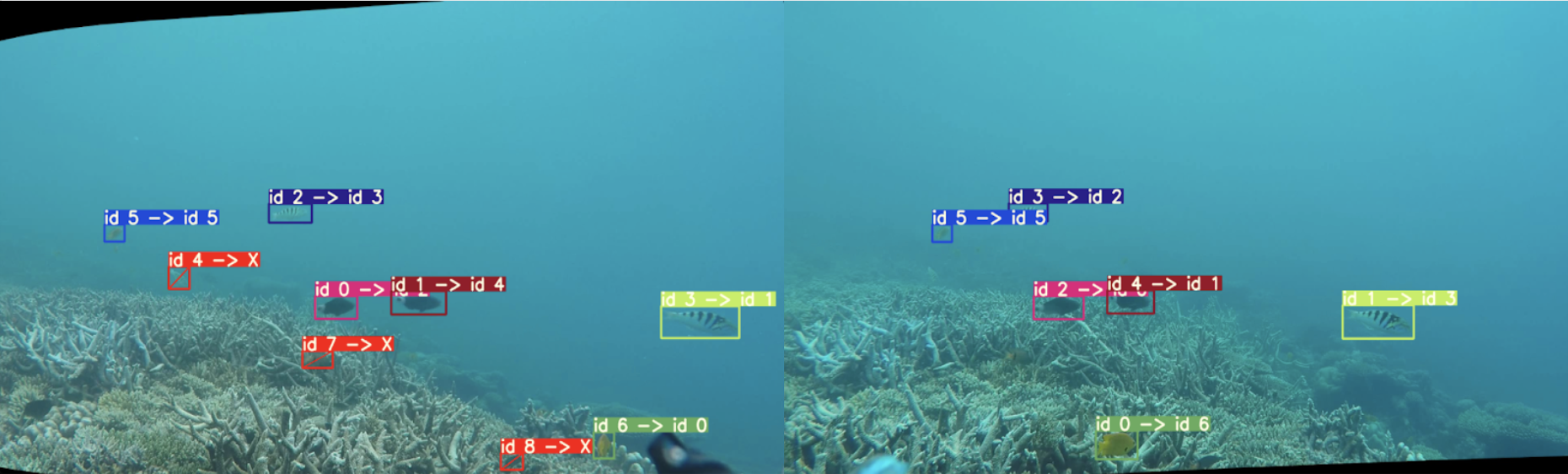}%
    \caption{Left and right frame highlighting paired BBs with the same color and complementary ID. Red-colored BBs indicate that there is no correspondence.}%
    \label{pairing}
\end{figure}

As shown in Figure~\ref{pairing}, each paired BB share the same color, and their Id is complementary. The red-colored BBs mean that we could not find any stereo correspondence. 
Then, we need to segment all pixels that belong to the fish to determine the snout and tail positions. 

Then, we need to segment all pixels that belong to the fish to determine the snout and tail positions. 

%
%
%
%
%
%

\subsection{Segmentation}

The segmentation allows for finding a binary mask of the fishes in order to determine their contours. Like detection and classification, a semantic image segmentation network, such as Mask-RCNN \cite{maskrcnn}, EfficientNet-L2 \cite{EffL2} or GoogleDeepLabv3+ \cite{DBLP:journals/corr/ChenPSA17},  has to be trained with examples. We chose the GoogleDeepLabv3+ implementation, which performs well on many tasks.\\
As segmentation takes some time and is unnecessary when a fish is not paired, hence, we only segmented paired fish to reduce the computation time.\\
Once a fish is paired on both I\textsubscript{L} and I\textsubscript{R}, their BB is cut out from the images. Then, these two cutouts pass through the segmentation network, which gives two binary masks, M\textsubscript{L} and M\textsubscript{R}, for the same fish detected on I\textsubscript{L} and I\textsubscript{R}.\\
Given that no fish segmentation was available, we created a small database. \\

\subsubsection*{FISH SEGMENTATION DATABASE}

From B\textsubscript{id}, we built our fish segmentation database B\textsubscript{seg} by randomly selecting 25 thumbnails per species for 1,600 thumbnails. First, we enlarged each thumbnail and manually drew a polygon on the fish edges. Next, the previously drawn mask is rescaled to the original thumbnail size. \\
Considering that our B\textsubscript{seg} is relatively small, we segment fish-looking objects rather than species.
Finally, we split B\textsubscript{seg} into three different datasets respecting the same ratio as we did for detection and classification. 

\subsubsection*{METHOD AND PARAMETERS}

\begin{figure}[!ht]
    \centering
    \includegraphics[width=10cm]{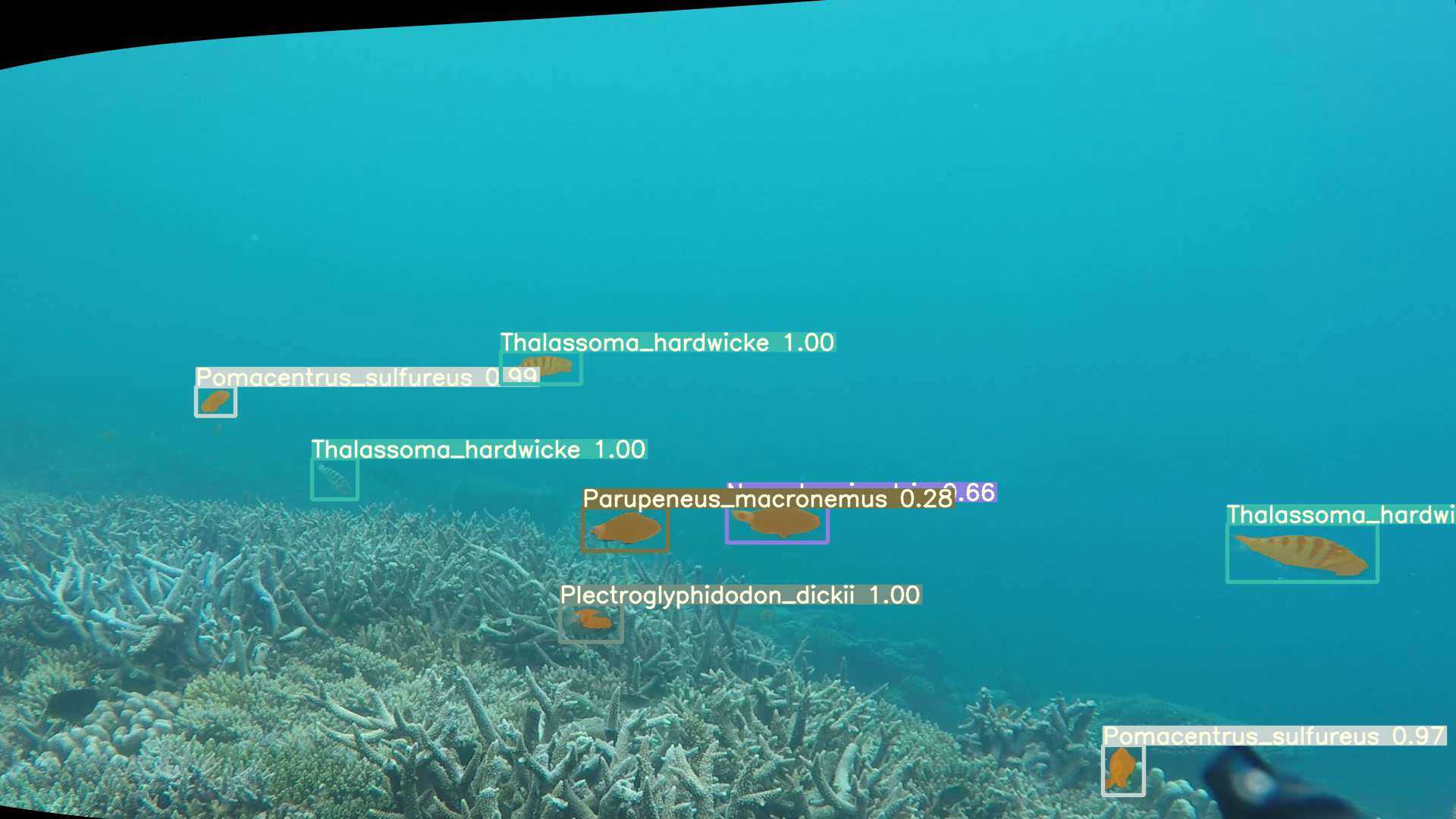}
    \caption{Shows the fish detection, identification, and, when BBs are paired, segmentation.}%
    \label{seg}
\end{figure}

As mentioned, we selected the DeepLabv3+ model with a ResNet-101 \cite{resnet} backbone pre-trained on ImageNet. We used the same parameters from the original paper: SGD optimizer; batch norm momentum 0.9; weight decay $10^{-5}$
; initial learning rate $10^{-4}$
, a learning rate policy where the learning rate is multiplied by $(1 - \frac{iter}{maxiter})^{0.9}$; the cross entropy loss function; and a batch size of 8. Additionally, we used an Early Stopping for the same reasons explained in the classification part. Finally, we obtained a mIoU of 84.4\% on the test set. \\
The result of the segmentation is shown in Figure~\ref{seg}.
 
%
%
%
%
%
%

\subsection{3D localization and measurements}

\begin{figure}[!htbp]
    \centering
    \includegraphics[width=12cm]{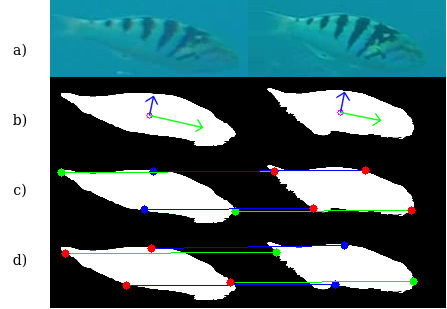}%
    \caption{Measurements points determination: a) Paired left and right fish; b) PCA on M\textsubscript{L} and M\textsubscript{R}; c) Projection of the set P\textsubscript{L}, represented in green and blue dots, on M\textsubscript{R} to obtain left to right corresponding points represented in red.; d) Projection of the set P\textsubscript{R}, represented in blue and red dots, on M\textsubscript{L} to obtain right to left corresponding points represented in red.}%
    \label{pca}
\end{figure}

To establish the snout and the tail, but also the bottom and the upper part positions from the fish, we first find the extreme points of M\textsubscript{L} and M\textsubscript{R}, namely the sets P\textsubscript{L} and P\textsubscript{R},  by determining the principal axis of inertia using a PCA, as shown in Figure~\ref{pca}. b). \\
Then, we project the set P\textsubscript{L} on M\textsubscript{R} using epipolar constraint, within a 3 pixels delta on the \emph{y} axis, to find left to right correspondences P\textsubscript{LR}, as shown in Figure~\ref{pca}. c). 
Given that M\textsubscript{L} and M\textsubscript{R} could be different, we repeat the same process to find right-to-left correspondences P\textsubscript{RL} from M\textsubscript{R} to M\textsubscript{L}, as shown in Figure~\ref{pca}. d).

\begin{figure}[!ht]
    \centering
    \includegraphics[width=10cm]{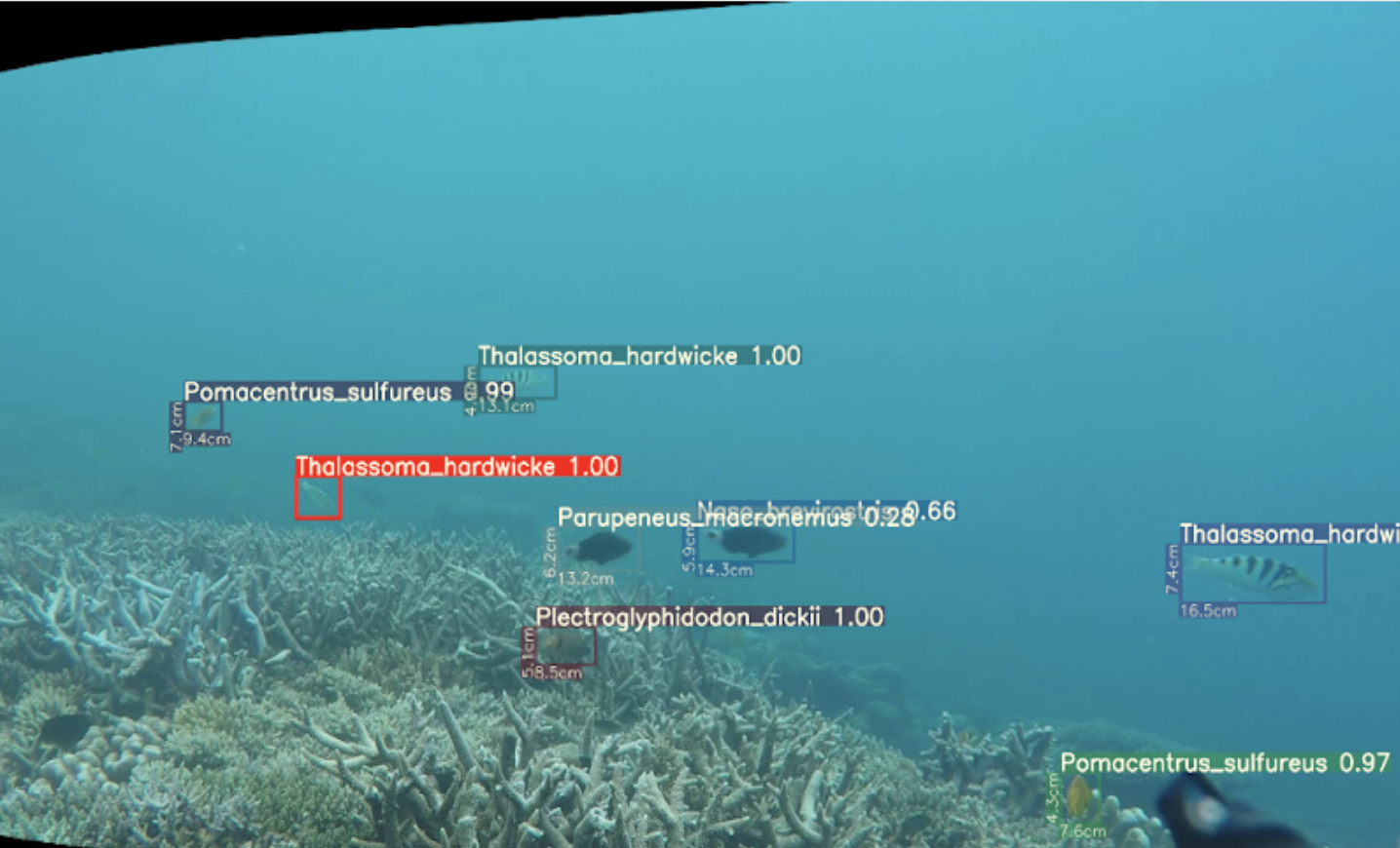}%
    \caption{Shows the fish detection, identification, and measurement. }%
    \label{measurement}
\end{figure}

Finally, we compute 3D point coordinates by triangulation using the projection matrices of the cameras for P\textsubscript{LR} and P\textsubscript{RL}. Therefore, we estimate the Fork length of the fish and its height for P\textsubscript{LR} and P\textsubscript{RL} by computing the Euclidean distances.\\
Because of the direction, we have two measurements for the length and the height. Therefore, we chose to take the mean of the lengths and the heights. \\
Additionally, we compute the barycenter for the sets P\textsubscript{L} and P\textsubscript{R}, respectively, the points B\textsubscript{L} and B\textsubscript{R}.
As we did for the length and the height, we find the left-to-right barycenter correspondence set B\textsubscript{LR} by taking the point B\textsubscript{L} and creating its correspondent point as B\textsubscript{R'} = (B\textsubscript{R\textsubscript{x}}, B\textsubscript{L\textsubscript{y}}).\\
Moreover, we find the right to left correspondence B\textsubscript{RL} = \{B\textsubscript{R}; B\textsubscript{L'}=(B\textsubscript{L\textsubscript{x}}, B\textsubscript{R\textsubscript{y}})\}.
Finally, we compute the 3D coordinates of the center for the sets B\textsubscript{LR} and B\textsubscript{RL} and take the mean between those two points.
Figure~\ref{measurement} shows the result of the measurement. 

%
%
%
%
%
%

\subsection{Tracking}

Ultimately, we added Multi-Object tracking to the pipeline to make it more robust. This addition enables us to track every BB over time, even though some detections miss. 
We used DeepSORT \cite{wojke_simple_2017}, based on SORT \cite{sort} to keep track of every detected fish on V\textsubscript{L}, and V\textsubscript{R}.\\
For each BB\textsubscript{L\textsubscript{i}}\textsuperscript{t\textsubscript{n}} on I\textsubscript{L}\textsuperscript{t\textsubscript{n}}, a new tracker is made, and is set as tentative.
The tracker is enabled when the IoU between BB\textsubscript{L\textsubscript{i}}\textsuperscript{t\textsubscript{n}} and BB\textsubscript{L\textsubscript{i'}}\textsuperscript{t\textsubscript{n+1}} is greater than a threshold for n = \{0; 1; 2\}.\\
Secondly, the motion model is built according to a standard Kalman filter with constant velocity and a linear observation model where the BB coordinates and aspect ratio are taken as direct observation of the object state. 
DeepSORT uses the squared Mahalanobis distance between predicted Kalman states and newly arrived measurements to assign a new BB to an existing tracker.
However, the Mahalanobis distance is a uniform metric for tracking through occlusions.\\
Therefore, a second metric is integrated into the assignment problem. Finally, DeepSORT keeps a gallery of each track's last 100 associated appearance descriptors.
We set this appearance descriptor to the feature vector we extracted from the classifier. 
Then, the second metric measures the smallest cosine distance between the BB\textsubscript{L\textsubscript{i}}\textsuperscript{t\textsubscript{n}} and the whole gallery.\\
Next, DeepSORT builds the association problem by combining metrics with a weighted sum and solves the problem by introducing a matching cascade considering the association likelihood.\\
Finally, we repeat the process for each BB\textsubscript{R\textsubscript{j}}\textsuperscript{t\textsubscript{n}} on I\textsubscript{R}\textsuperscript{t\textsubscript{n}}, allowing us to track every classification and, when possible, measurements for the same fish through its appearance in the video.

%
%
%
%
%
%

\subsection{Fusion and refinement}

While a fish can be identified and measured automatically, errors could occur through classification and segmentation. To alleviate those issues, we chose to take advantage of the tracking to smooth the results.\\
First, we associate each left track with its corresponding right track, considering that both left and right fish should be paired in the same time frame. Therefore, we have a unique track for recording left and right classifications and measurements. \\
Secondly, we sum up the classification top-5 score each time a fish is identified and retain the most likely identification for each unique track.\\ 
Additionally, we chose to take the median of all measurements for the length and height. 

\section{Results}
Our method yields a multifaceted range of results due to its comprehensive capabilities. The initial 2D positional localization attained through the detection module serves as a foundation for temporal fish enumeration. Furthermore, by incorporating species identification via the classification module, we achieve a comprehensive assessment of piscine diversity.

The segmentation module, when applied, facilitates an in-depth analysis of the morphological characteristics of various species. It allows for an examination of species-specific shapes and structures, thus offering a more nuanced understanding of the aquatic fauna. Additionally, the integration of 3D positional localization, coupled with segmentation data, enables accurate estimations of fish sizes. This information, coupled with species identification and population dynamics over time, allows for the calculation of biomass within the ecosystem.

Furthermore, our method’s ability to track individual fish across a video sequence enables a thorough analysis of their kinematic aspects. This includes the speed, acceleration profiles, and trajectories, visualized in Figure~\ref{chart}. Additionally, the system provides opportunities to study intra-species interactions and behavioral patterns.

\begin{figure}[!htbp]
    \centering
    \includegraphics[width=8cm]{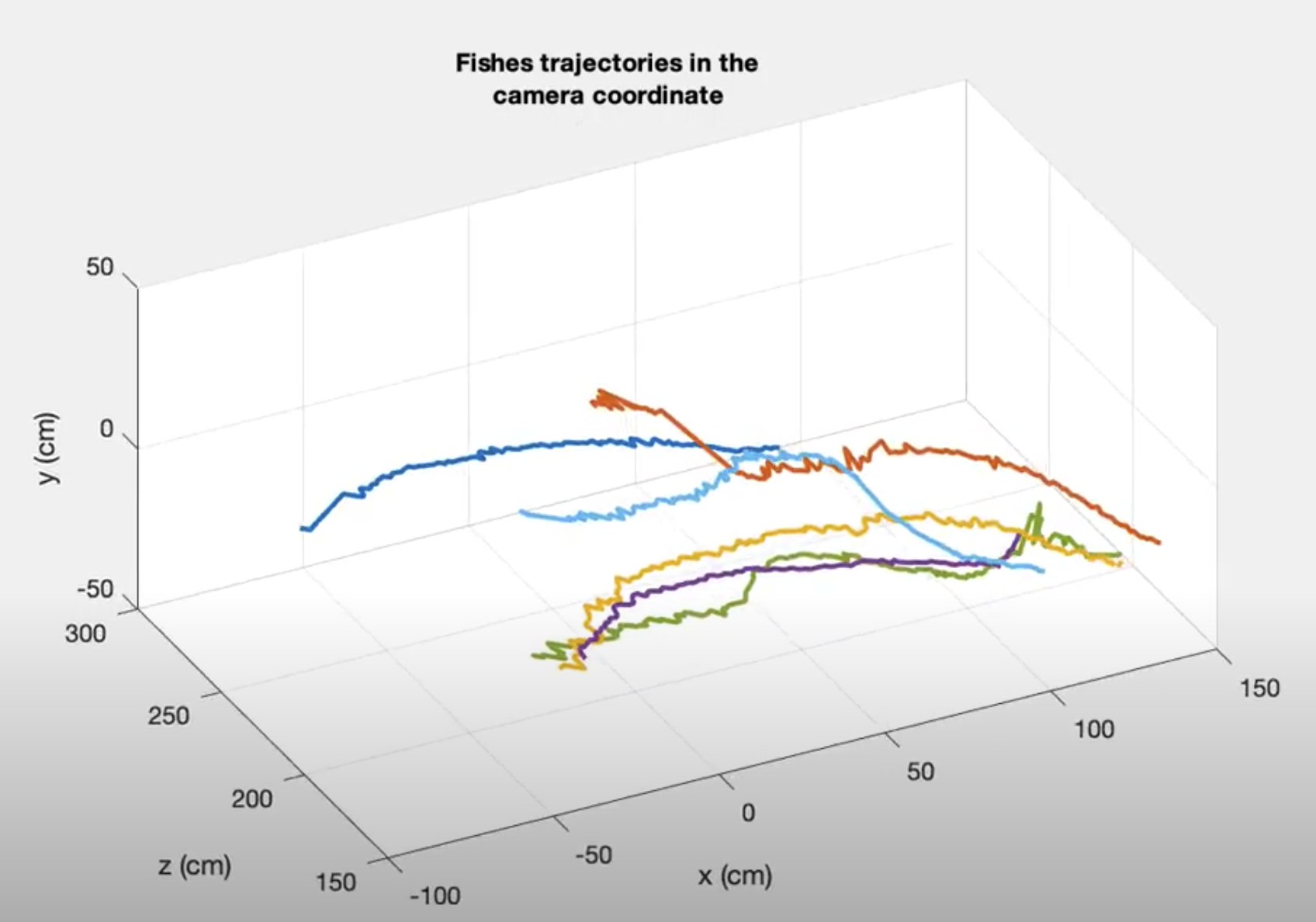}%
    \caption{\textbf{3D trajectories} Shows captured 3D fish trajectories for a short period of time, where each color represents a different fish.}%
    \label{chart}
\end{figure}

\section{Discussion}
While we present an automatic fish measurement and identification pipeline, every block could be improved separately.\\
As detection is challenging in underwater images, we could augment B\textsubscript{det} with new examples. However, since annotation is time-consuming, we could use recent semi-supervised learning techniques \cite{semi-sup}, \cite{Sohn2020ASS}, and \cite{liu2021unbiased}. 
While having just a few annotated examples, we have thousands of underwater video records. The semi-supervised learning approach creates new examples from unlabeled data and is proven to improve generalization.\\
Given that the classifier will always classify whether or not the input is part of the database, the unseen classification object is an open issue. We could set a threshold to class a classification from correct to uncertain \cite{villon_open_nodate}.\\
While the mIoU seems conclusive on the test set for the segmentation, the generalization still needs to be proven. 
We could build a more extensive database to improve the segmentation and use weakly supervised techniques. For instance, Puzzle-CAM \cite{puzzle} uses the class activation map of the classification network with a puzzle method to create new examples. 
Weakly supervised learning could make use of the whole B\textsubscript{id}, and therefore, improve the generalization.\\
While PCA is a decent method to find the snout and the tail position, it requires excellent fish segmentation. Moreover, even though the segmentation is good enough, some fish could be taller than long. In that case, we could use a CNN to determine those positions similarly to DeepFish \cite{saleh_realistic_2020}, which finds the fish's center position using points annotation examples.
The tracking component could also be improved by using recent online tracking as STGT \cite{chu2021transmot} or by using offline tracking methods \cite{offline1} and \cite{offline2}. Nevertheless, since we work on stereo images, we could also integrate the 3D information into the tracking \cite{li2019stereo}.


\section*{Acknowledgments}
This work was publicly funded through ANR (the French National Research Agency) under the "Investissements d'avenir" program (PIA) with the reference ANR-16-IDEX-0006

\section*{Authors' contributions}
All authors conceived the ideas and designed the methodology; J.S.G. led the writing of the manuscript. All authors contributed critically to the drafts and gave final approval for publication.

\section*{Conflict of Interest statement}
Authors have no conflict of interest to declare. 

\bibliographystyle{IEEEbib}

\bibliography{refs}

\section*{Supplementary Materials}
\begin{longtable}{|l|r|}
\hline
\textbf{Species} & \textbf{Number of thumbnails} \\
\hline
\endhead
Abudefduf & 5 \\
\hline
Abudefduf sexfasciatus & 27 \\
\hline
Abudefduf sparoides & 102 \\
\hline
Abudefduf vaigiensis & 62 \\
\hline
Acanthurus & 86 \\
\hline
Acanthurus dussumieri & 172 \\
\hline
Acanthurus fowleri & 7 \\
\hline
Acanthurus leucosternon & 55 \\
\hline
Acanthurus nigricans & 11 \\
\hline
Acanthurus thompsoni & 285 \\
\hline
Aethaloperca rogaa & 99 \\
\hline
Aethaloperca rogaa juv & 7 \\
\hline
Amanses scopas & 58 \\
\hline
Amblyglyphidodon indicus & 33 \\
\hline
Amphiprion & 46 \\
\hline
Amphiprion latifasciatus & 40 \\
\hline
Amphiprion sandaracinos & 49 \\
\hline
Anampses twistii & 16 \\
\hline
Aphareus furca & 17 \\
\hline
Apogonidae & 8 \\
\hline
Apolemichthys trimaculatus & 5 \\
\hline
Arothron & 97 \\
\hline
Arothron nigropunctatus & 30 \\
\hline
Balistapus & 8 \\
\hline
Balistapus undulatus & 109 \\
\hline
Balistidae & 5 \\
\hline
Bodianus anthioides & 8 \\
\hline
Bodianus axillaris juv & 7 \\
\hline
Bodianus diana & 11 \\
\hline
Caesio lunaris & 39 \\
\hline
Caesio teres & 179 \\
\hline
Canthigaster valentini & 11 \\
\hline
Carangidae & 28 \\
\hline
Carangoides & 4 \\
\hline
Caranx melampygus & 5 \\
\hline
Carcharhinus melanopterus & 7 \\
\hline
Centropyge & 5 \\
\hline
Centropyge multispinis & 83 \\
\hline
Cephalopholis argus & 61 \\
\hline
Cephalopholis juv & 2 \\
\hline
Cetoscarus ocellatus & 6 \\
\hline
Chaetodon & 35 \\
\hline
Chaetodon auriga & 92 \\
\hline
Chaetodon falcula & 42 \\
\hline
Chaetodon guttatissimus & 233 \\
\hline
Chaetodon kleinii & 44 \\
\hline
Chaetodon lunula & 10 \\
\hline
Chaetodon lunulatus & 3 \\
\hline
Chaetodon meyeri & 63 \\
\hline
Chaetodon striatus & 82 \\
\hline
Naso & 3 \\
\hline
Naso elegans & 62 \\
\hline
Nemanthias carberryi female & 8 \\
\hline
Nemanthias carberryi male & 71 \\
\hline
Nemateleotris magnifica & 252 \\
\hline
Novaculichthys taeniourus & 2 \\
\hline
Ostracion meleagris female & 5 \\
\hline
Oxymonacanthus longirostris & 68 \\
\hline
Paracirrhites forsteri & 148 \\
\hline
Parupeneus macronemus & 3 \\
\hline
Parupeneus trifasciatus & 11 \\
\hline
Plectorhinchus obscurus & 40 \\
\hline
Plectroglyphidodon dickii & 229 \\
\hline
Plectroglyphidodon johnstonianus & 12 \\
\hline
Plectroglyphidodon lacrymatus & 105 \\
\hline
Plectropomus laevis & 11 \\
\hline
Plectropomus punctatus & 53 \\
\hline
Pomacentrus & 10 \\
\hline
Pomacentrus caeruleus & 14 \\
\hline
Pseudanthias & 628 \\
\hline
Pseudanthias bicolor female & 85 \\
\hline
Pseudanthias squamipinnis female & 843 \\
\hline
Pseudanthias squamipinnis male & 384 \\
\hline
Pseudocheilinus hexataenia & 23 \\
\hline
Ptereleotris evides & 30 \\
\hline
Pterocaesio & 4 \\
\hline
Pterocaesio tile & 174 \\
\hline
Pygoplites diacanthus & 148 \\
\hline
Sargocentron & 6 \\
\hline
Sargocentron caudimaculatum & 8 \\
\hline
Sargocentron cornutum & 13 \\
\hline
Scaridae & 41 \\
\hline
Scarus & 28 \\
\hline
Scarus frenatus male & 36 \\
\hline
Scarus rubroviolaceus male & 19 \\
\hline
Scarus sordidus female & 23 \\
\hline
Serranidae & 11 \\
\hline
Stegastes & 7 \\
\hline
Stegastes pelicieri & 28 \\
\hline
Sufflamen bursa & 22 \\
\hline
Thalassoma & 27 \\
\hline
Thalassoma amblycephalum & 30 \\
\hline
Thalassoma amblycephalum juv & 114 \\
\hline
Thalassoma hardwicke & 410 \\
\hline
Thalassoma hardwicke female & 1 \\
\hline
Thalassoma hardwicke male & 21 \\
\hline
Thalassoma hebraicum & 72 \\
\hline
Thalassoma lunare & 77 \\
\hline
Variola louti & 16 \\
\hline
Zanclus cornutus & 187 \\
\hline
Chaetodon trifascialis & 56 \\
\hline
Chaetodon trifasciatus & 268 \\
\hline
Chaetodon ulietensis & 14 \\
\hline
Chaetodontidae & 4 \\
\hline
Chromis & 152 \\
\hline
Chromis atripectoralis OR Chromis viridis & 5 \\
\hline
Chromis lepidolepis & 199 \\
\hline
Chromis viridis & 765 \\
\hline
Chromis xanthura & 36 \\
\hline
Ctenochaetus & 22 \\
\hline
Ctenochaetus strigosus & 2 \\
\hline
Ctenochaetus truncatus & 427 \\
\hline
Ecsenius midas & 28 \\
\hline
Epinephelus & 4 \\
\hline
Epinephelus fasciatus & 7 \\
\hline
Epinephelus flavocaeruleus & 11 \\
\hline
Epinephelus rivulatus & 7 \\
\hline
Epinephelus tukula & 5 \\
\hline
Fistularia commersonii & 10 \\
\hline
Gnathodentex & 16 \\
\hline
Gnathodentex aureolineatus & 122 \\
\hline
Gomphosus caeruleus female & 259 \\
\hline
Gomphosus caeruleus male & 272 \\
\hline
Gracila albomarginata & 22 \\
\hline
Gunnellichthys monostigma & 10 \\
\hline
Halichoeres & 13 \\
\hline
Halichoeres hortulanus & 67 \\
\hline
Halichoeres hortulanus transitional & 29 \\
\hline
Hemigymnus fasciatus & 98 \\
\hline
Heniochus & 16 \\
\hline
Hologymnosus annulatus & 11 \\
\hline
Kyphosus & 18 \\
\hline
Kyphosus cinerascens & 9 \\
\hline
Kyphosus vaigiensis & 1 \\
\hline
Labrichthys unilineatus & 19 \\
\hline
Labridae & 144 \\
\hline
Labroides & 7 \\
\hline
Labroides dimidiatus & 167 \\
\hline
Lutjanus & 11 \\
\hline
Lutjanus bohar & 11 \\
\hline
Lutjanus gibbus & 84 \\
\hline
Macolor niger juv & 8 \\
\hline
Malacanthidae & 2 \\
\hline
Monotaxis grandoculis & 30 \\
\hline
Myripristis & 20 \\
\hline
Myripristis kuntee & 68 \\
\hline
Myripristis murdjan & 11 \\
\hline
Zebrasoma & 15 \\
\hline
Zebrasoma rostratum & 283 \\
\hline
Zebrasoma scopas & 231 \\
\hline
unknown fish & 91 \\
\hline

\caption{\textbf{B\textsubscript{det} – 3367 frames}}
\label{Bdet}
\end{longtable}
\begin{longtable}{|l|r|}
\hline
\textbf{Species} & \textbf{Number of thumbnails} \\
\hline
\endhead
Abudefduf sparoides & 3,068 \\
\hline
Abudefduf vaigiensis & 5,124 \\
\hline
Acanthurus leucosternon & 3,259 \\
\hline
Acanthurus lineatus & 1,008 \\
\hline
Acanthurus tennenti & 1,115 \\
\hline
Acanthurus thompsoni & 3,201 \\
\hline
Aethaloperca rogaa & 1,885 \\
\hline
Amblyglyphidodon indicus & 1,188 \\
\hline
Balistapus undulatus & 1,787 \\
\hline
Caesio lunaris & 1,147 \\
\hline
Caesio xanthonota & 2,817 \\
\hline
Cephalopholis argus & 1,991 \\
\hline
Chaetodon auriga & 2,134 \\
\hline
Chaetodon guttatissimus & 1,182 \\
\hline
Chaetodon kleinii & 1,378 \\
\hline
Chaetodon trifascialis & 5,234 \\
\hline
Chaetodon trifasciatus & 4,421 \\
\hline
Cheilinus fasciatus & 1,333 \\
\hline
Chromis dimidiata & 7,004 \\
\hline
Chromis opercularis & 1,525 \\
\hline
Chromis ternatensis & 3,640 \\
\hline
Chromis viridis & 1,425 \\
\hline
Chromis weberi & 4,977 \\
\hline
Dascyllus trimaculatus & 1,103 \\
\hline
Gnathodentex aureolineatus & 1,796 \\
\hline
Gomphosus caeruleus female & 3,178 \\
\hline
Gomphosus caeruleus male & 3,131 \\
\hline
Halichoeres hortulanus & 3,192 \\
\hline
Hemigymnus fasciatus & 2,285 \\
\hline
Hemitaurichthys zoster & 1,136 \\
\hline
Labroides dimidiatus & 2,329 \\
\hline
Lutjanus bohar & 1,246 \\
\hline
Lutjanus bohar juv & 1,588 \\
\hline
Macolor niger & 2,643 \\
\hline
Macolor niger juv & 1,334 \\
\hline
Monotaxis grandoculis & 2,045 \\
\hline
Monotaxis grandoculis juv & 1,848 \\
\hline
Naso brevirostris & 1,134 \\
\hline
Naso elegans & 7,345 \\
\hline
Naso vlamingii & 1,684 \\
\hline
Odonus niger & 1,168 \\
\hline
Oxymonacanthus longirostris & 2,553 \\
\hline
Parupeneus macronemus & 3,469 \\
\hline
Plectroglyphidodon dickii & 1,458 \\
\hline
Plectroglyphidodon lacrymatus & 2,497 \\
\hline
Pomacanthus imperator & 1,847 \\
\hline
Pomacentrus sulfureus & 5,409 \\
\hline
Pseudanthias squamipinnis female & 1,542 \\
\hline
Pygoplites diacanthus & 3,746 \\
\hline
Sargocentron caudimaculatum & 1,111 \\
\hline
Sargocentron diadema & 1,031 \\
\hline
Scarus tricolor female & 1,140 \\
\hline
Thalassoma hardwicke & 4,951 \\
\hline
Thalassoma hebraicum & 2,173 \\
\hline
Zanclus cornutus & 3,876 \\
\hline
Zebrasoma scopas & 4,970 \\
\hline
 calcareous & 7,323 \\
\hline
 corals hard & 9,500 \\
\hline
 corals soft & 1,948 \\
\hline
 dark & 5,068 \\
\hline
 macroalgae & 1,730 \\
\hline
 sand & 2,066 \\
\hline
water blue & 1,000 \\
\hline
water surface & 4,050 \\
\hline
\caption{\textbf{B\textsubscript{id} – 176,486 thumbnails}}
\label{Bid}
\end{longtable}

\end{document}